\documentclass{article}

\usepackage[nonatbib,preprint]{neurips_2022}
\usepackage[numbers]{natbib}
\usepackage[utf8]{inputenc}
\usepackage[T1]{fontenc}
\usepackage{hyperref}
\usepackage{url}
\usepackage{booktabs}
\usepackage{amsfonts}
\usepackage{nicefrac}
\usepackage{microtype}
\usepackage{xcolor}

\usepackage{mathtools}

\usepackage{mdframed}

\title{Early Transferability of Adversarial Examples in Deep Neural Networks}
\author{ Oriel BenShmuel 
\\
Faculty of Math\&CS \\
Weizmann Institute of Science \\
Israel \\
\texttt{oriel.benshmuel@weizmann.ac.il}
}
\date{\vspace{-5ex}}

\begin{document}

\date{\vspace{-5ex}}

\maketitle

\begin{abstract}
This paper will describe and analyze a new phenomenon that was not known before, which we call ``Early Transferability''. Its essence is that the adversarial perturbations transfer among different networks even at extremely early stages in their training. In fact, one can initialize two networks with two different independent choices of random weights and measure the angle between their adversarial perturbations after each step of the training. What we discovered was that these two adversarial directions started to align with each other already after the first few training steps (which typically use only a small fraction of the available training data), even though the accuracy of the two networks hadn't started to improve from their initial bad values due to the early stage of the training. The purpose of this paper is to present this phenomenon experimentally and propose plausible explanations for some of its properties.

\end{abstract}

\section{Introduction}

Previous findings had demonstrated the surprising
fact that tiny adversarial perturbations could fool trained deep neural networks \cite{szegedy2013intriguing,biggio2013evasion}. The distorted images, known as ``adversarial examples'' have become a major concern for machine learning practitioners since they enable attackers to fool even the best trained deep neural networks by slightly modifying the input in a way that is not perceptible to humans. This fact naturally raised the curiosity of the deep learning community \cite{liu2017delving,ilyas2019adversarial,kurakin2018adversarial,carlini2017adversarial,xu2017feature,athalye2018obfuscated,grosse2017adversarial,carlini2018audio}. In the same paper \cite{szegedy2013intriguing}, Szegedy et al. also mentioned the ``cross-generalization'' property, where the same adversarial example can usually fool a wide variety of networks, which may differ in architecture and even disjoint datasets. This poorly understood phenomenon is called ``Transferability’’.

Shortly afterward, it was discovered that different models with similar decision boundaries enable transferability \cite{tramer2017space}. In fact, a strong correlation between adversarial directions of the two models (concerning the same sample) implies transferability (but not necessarily vice versa). Generally, a strong correlation between two random vectors is very unlikely due to the high dimension of the input space (accurate measurements appear in Appendix \ref{appendix:angles_random_vectors}).

Black-box adversarial attacks rely on transferability \cite{liu2016delving}. By attacking a surrogate model that the attacker trained, we get adversarial examples that could fool the original network with high probability (without further information concerning the parameters of the original model itself).

Studied discuss the nature of transferability by estimating the space of adversarial examples \cite{tramer2017space} or alternatively introducing new mental images that aim to explain the phenomenon, such as robust and non-robust features \cite{ilyas2019adversarial} or the linear nature of neural networks \cite{goodfellow2015explaining,guo2020backpropagating}. Another series of studies discuss defense methods \cite{madry2017towards,Liao_2018_CVPR,10.1145/3128572.3140449,Ru2020BayesOpt,NEURIPS2019_7503cfac,ho2020contrastive} and attack methods \cite{8294186,Dong_2018_CVPR,pmlr-v97-guo19a,NEURIPS2019_32508f53,pmlr-v80-dai18b,tramer2020adaptive}.

The adversarial perturbations of the same image, computed by using two models with random weights (before the training phase), are random as well and tend to be perpendicular (as explained in Appendix \ref{appendix:angles_random_vectors} and observed in sections \ref{experimental_results}, \ref{long_term}, and Appendix \ref{appendix:additional_results}). A reasonable assumption might be that transferability occurs after the networks learn some features of the data, which usually takes several epochs in non-trivial datasets.

This paper will describe and explain a new phenomenon that was not known before. Its occurrence influences and emphasizes the transferability effect at the very beginning of the learning process with possible long-term consequences. We call this occurrence ``Early Transferability'', and although it is characterized by the same behavior as the general notion of transferability as defined above, it is caused by reasons that are relevant mainly during the early stages of the training phase and has additional and counter-intuitive properties. First, we describe the new phenomenon and its counter-intuitive properties with an extensive explanation of the goal of presenting it. Then, we experimentally present the occurrence using multiple different settings. Finally, we propose a partial explanation for the occurrence while we try to understand the impact of each network's parameters concerning the new phenomenon.

\section{Early transferability}

Even though the transferability occurrence is still vague, typically, we would expect it to happen as part of the process of learning the data. In particular, it could only occur after some properties of the data were learned, which usually takes a few epochs in non-trivial datasets (as a function of the data complexity). This is not the case for early transferability.

\subsection{Early transferability and complex datasets}
Naturally, complex datasets require advanced architectures and more epochs for the network to perform well, as opposed to basic datasets. Assuming the general notion of transferability addresses trained networks, we expect the effect to evolve as part of the network's learning process. This assumption points to a difference between the transferability evolution using complex datasets and basic datasets.

In contrast, the early transferability effect has no relation to the ability of the network to learn and generalize. It might happen even \emph{before} the network started to improve its initial $50\%$ accuracy, or even in the first \emph{step} (which is based on a tiny subset of the data as a function of the batch size), rather than after several \emph{epochs}. As a result, early transferability might occur even when the data is highly complex, the current architecture could not learn it, or even if the data is not learnable at all (for example, there is no semantic difference between classes, or alternatively, the data is built from random noise).

\subsection{Long-term effect}
As mentioned before, the impact of early transferability might differ as a function of the train settings. Even though early transferability occurs in the early stages of the training, it might influence the long-term training and the long-term transferability due to the initial impact of the effect (for further details, see section \ref{long_term}).

\subsection{Controlling early transferability}
This paper will explain and experimentally demonstrate the influence of early transferability for different types of hyper-parameters. Understanding its properties and behavior could imply the parameters that influence it. Controlling these parameters helps us emphasize or reduce this effect or generally be aware of it.

\subsection{The goal of early transferability}
\label{goal}
Early transferability has three main goals:
\begin{enumerate}
\item \emph{Prevent confusion:}
Showing this phenomenon aims to prevent confusion between traditional transferability and early transferability. Being aware of early transferability might shed light on various fields of research regarding transferability. Studies that discuss transferability could reach better conclusions when isolating the traditional transferability effect. For example, when investigating the relationship between transferability and complex datasets or the evolution of this occurrence during training.

\item \emph{Reliability:} When studying transferability, conclusions might include biased results toward early transferability rather than the traditional transferability. It is essential to address it since the effect of early transferability can be controlled by the input hyper-parameters and, therefore, could create a massive impact on the results (in the long term as well). We consider the early transferability effect as a nonobjective effect, which was influenced by the trainer. We would address the traditional transferability as the objective effect that evolves naturally and independently of the inserted parameters. 

Moreover, black-box attacks are more reliable when based on traditional transferability rather than early transferability since early transferability assumes emphasized effect using some set of parameters. In contrast, using different parameters might cause a weaker effect. In the process of demonstrating black-box attacks, we might choose the reference model ourselves. At the same time, our experimental results and conclusions might be biased due to a more substantial early transferability effect rather than a different choice of parameters that could massively weaken the presented results.

\item \emph{What to do?} Being aware of this effect might lead to a better understanding of how transferability behaves. However, more importantly, understanding it helps to reduce (or emphasize) the effect in order to achieve better conclusions (or generate new forms of attacks).
\end{enumerate}

\section{Experimental results}
\label{experimental_results}
In this section, we present the ``Early Transferability'' effect by generating quantitative experiments based on a wide range of settings.

\subsection{Framework}
The following properties in the analysis will help us witness the transferability in the very early stages of the training:
\begin{enumerate}
\item \emph{Steps:} Early transferability refers to the type of transferability that happens already in the early stages of the training. In order to test that, the experiments will use ``zoom-in'' observations of the training process, such that the provided details are shown in scale of steps (where each optimizer step includes only one batch from the training set) rather than in scale of epochs (where each epoch includes the entire train set).

\item \emph{Angles:} For each step, we compute the angle between the adversarial directions of the two models for each sample. The displayed result is an average of over $100$ samples from the \emph{test set} (to emphasize that this kind of transferability has nothing to do with the specific train images that the model observed). We are interested in deviations from the $90^{\circ}$ angle expected for pairs of random vectors. In some cases, the correlation was negative. Therefore, for each angle $a>90^{\circ}$, we will display $180^{\circ}-a$.
\item \emph{Accuracy:} One of the mentioned claims states that early transferability is unrelated to the accuracy of the model. Specifically, the model might improve its test accuracy on the data after every few steps, or (in more complex cases) it might stay steady (at $50\%$ accuracy) for many steps. In this set of experiments, we show that we might get an extremely high transferability effect, even in the cases where the model hasn't learned anything yet (in terms of test accuracy). Therefore we will display the test accuracy of each model.
\end{enumerate}

Generally, we will use Adam optimizer \cite{kingma2017adam}. The same effect is true for the RMSprop optimizer \cite{tieleman2012lecture}, and it also appears when using SGD \cite{spall2005introduction} (with and without momentum \cite{pmlr-v28-sutskever13}). Still, for SGD, the effect is smaller due to reasons that appear in subsection \ref{optimizers}. Additionally, we are using different shuffles for the dataset in each of the two models to show that the effect occurs even when we train the models on entirely \emph{different} images (expressed with disjoint batches) during the early steps. The experiments were executed using GPU ``Tesla K80''.

\subsection{Results}
\begin{figure}[htp!]
     \includegraphics[width=1\textwidth]{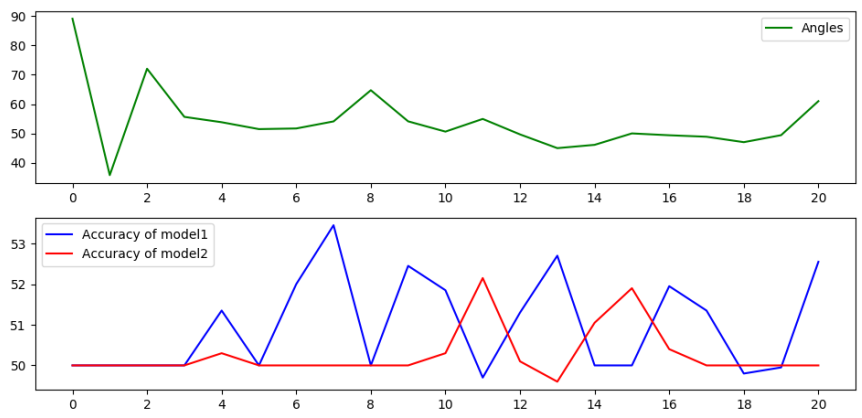}
     \caption{The ``Early Transferability'' effect by ``zooming in'' into the beginning of the learning process.}
     \label{fig:main_case}
\end{figure}

In Figure \ref{fig:main_case}, we can observe the results of a pair of deep fully-connected networks (according to the ``Deep network'' architecture mentioned in Appendix \ref{appendix:architectures}) trained to classify between \emph{Cat} and \emph{Dog} of the CIFAR10 dataset \cite{CIFAR10} with a learning rate of $1e-2$ and batch size of $128$ under the described analysis. In the graph, we can see that the starting point of the networks, when they are randomly initialized, produces an angle of nearly $90^{\circ}$ between the adversarial directions as should be in high dimensional space, while both of the accuracy values of the networks are $50\%$ as expected. Observing the same values after the first step shows us a drastic change in the angle to less than $40^{\circ}$, while the accuracy is still steady at $50\%$ (it will start improving only at steps $3$ and $10$ of the two networks, respectively).

As discussed earlier, an average angle between randomly sampled vectors (in high dimensions) should be very close to $90^{\circ}$. At the same time,  the standard deviation is small as a function of the space dimension (for more details, see Appendix \ref{appendix:angles_random_vectors}). Based on that, even minor deviations from the mean value are significant and extremely unlikely.

In the presented case, we use a small batch size ($128$) compared to the entire dataset and especially compared to the input space dimension, which is $3072$. This ratio leaves many degrees of freedom for the network to ``learn'' from the given data and change its weights accordingly, and yet the minimally trained networks are already transferable.

\noindent

\begin{figure}[htp!]
     \includegraphics[width=1\textwidth]{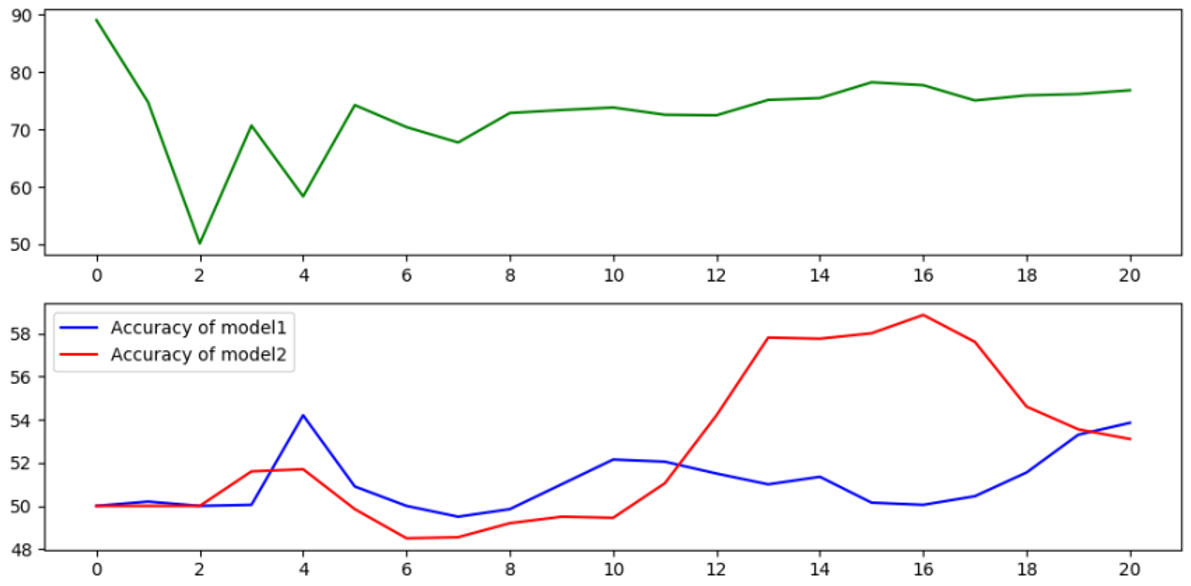}
     \caption{The ``Early Transferability'' effect using CNN.}
     \label{fig:main_case_conv}
\end{figure}

The effect of early transferability is not unique to fully-connected networks. The phenomenon also exists in convolutional networks. In Figure \ref{fig:main_case_conv}, we can observe the same analysis (with similar categories) using Architecture B (convolutional network, see Appendix \ref{appendix:architectures}), batch size of $128$, and learning rate of $1e-3$.

\begin{figure}[htp!]
     \includegraphics[width=1\textwidth]{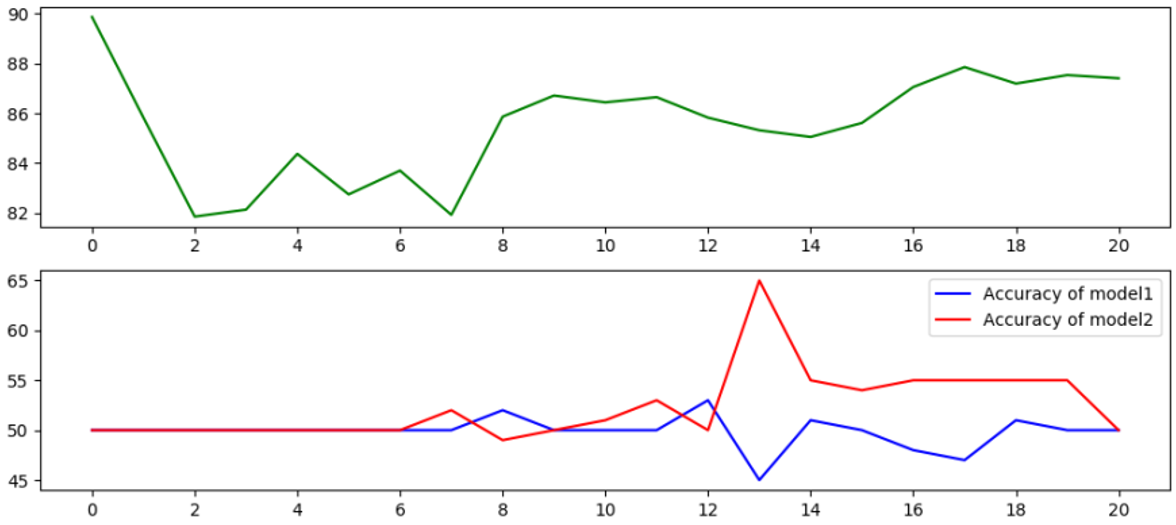}
     \caption{The ``Early Transferability'' effect using the vgg11 network trained with ImageNet.}
     \label{fig:main_case_imagenet}
\end{figure}

Similarly, the effect also occurs for complex datasets such as the ImageNet dataset \cite{IMAGENET}. In Figure \ref{fig:main_case_imagenet}, we present the described analysis using the VGG11 network \cite{Simonyan15} trained to classify \emph{Goldfish} and \emph{White Shark} from the ImageNet dataset, with a batch size of $16$ and a learning rate of $1e-3$.

For additional experiments with different settings, see Appendix \ref{appendix:additional_results}.

\section{The first steps of the training process}
\label{Occurrence explanation}
This section proposes a simple explanation based on some interesting results for the most basic case, fully-connected networks trained with SGD optimizer, which leads to a high-level understanding of early transferability. In particular, the models were trained with \emph{disjoint} batches (randomly sampled from the dataset). In subsection \ref{optimizers}, we will elaborate on optimizers that might emphasize this effect and the possible long-term influence (in subsection \ref{long_term}).

\subsection{Notation}
We denote by $\theta_l[j]$ the $j$'th row in the weights matrix $\theta_l$ from size $k \times n$ (where $0\leq j< k$) of layer number $l$ in the network. We also denote by $\partial \theta_l[j]$ the change in the weights $\theta_l[j]$ when calculating the gradient of the loss function with respect to the weights.

For the sake of simplicity, we usually address $\{\theta_l[j]\}_{0\leq j< k}$ and $\{\partial \theta_l[j]\}_{0\leq j< k}$ as ``vectors of weights'' and ``vectors of updates'', respectively.

\subsection{Discussion}
For a pair of randomly initialized networks, we get almost perpendicular adversarial directions (based on similar input). This angle between the two adversarial vectors is typical for the high-dimensional input space (as we explain in Appendix \ref{appendix:angles_random_vectors}). However, the experimental results point to correlated adversarial vectors, sometimes even after the first step. The only change we made in the initial settings is updating the weights of the networks. Following that, we can observe the weights of a fully-connected network as a set of vectors rather than a matrix of weights, such that each vector represents the relation between a neuron of the following layer and the entire set of weights connecting it to the current layer.

During the first steps of the training process, we encountered the following occurrences, which we experimentally support in this section:
\begin{enumerate}
    \item Focusing on the first layer of each network, where the ``vectors of updates'' of the first network are $A=\{\partial \theta_1[j]\}_{0\leq j< k}$ and the ``vectors of updates'' of the second network are $B=\{\partial \theta'_1[j']\}_{0\leq j'< k'}$, we get that each pair of vectors from the united set $A\cup B$ is highly correlated with high probability. In particular, it occurs even when using a different batch of images for each model and a non-linear objective.
    
    \item For a fully-connected network with a correlated set of ``vectors of weights'', we get that the adversarial direction is correlated to them as well. In fact, the adversarial direction will be correlated to each one of the vectors in the set with high probability.
    
\end{enumerate}

Combining these two properties together, a large enough impact (depending on the learning rate, batch size, and optimizer) of the ``vectors of updates'' will convert the initial vectors of weights to be correlated as well, and therefore, the networks will generate correlated adversarial directions with high probability.

Additionally, for a relatively ``slow'' optimizer (which might have a small impact on the weights update in the first steps), such as SGD, the effect is smaller and requires a higher learning rate to witness the effect. In contrast, for relatively ``aggressive'' optimizers, such as the Adam optimizer, the effect will be much more significant and will clearly appear even when using relatively small learning rates (more information, see subsection \ref{optimizers}).

Concluding all this new information (which we support experimentally), we get that for the right amount of impact of the weights update, the adversarial vectors are correlated with high probability.

\subsection{Experimental support}
For the fully-connected architectures mentioned in Appendix \ref{appendix:architectures}, the categories \emph{Cat} and \emph{Dog} of the dataset CIFAR10 \cite{CIFAR10}, the negative log-likelihood loss function, and for a single training step based on a \emph{different} batch of images for each model, we will check the correlations between $\partial \theta_1[j_1]$ and $\partial \theta_1[j_2]$ (for $0\leq j_1\neq j_2<k$) among each model separately (using only one model) and between two different models (for similar architectures and different architectures). For convenience, we will ignore the correlation signs by transforming angles $a>90^{\circ}$ to be $180^{\circ}-a$ since we are only interested in deviations from the expected $90^{\circ}$ angle. The weights of the first layer are $\theta_1$ of size $k\times n$ for the first model and $\theta'_1$ of size $k'\times n$ for the second model.

For a single step, batch size $30$, input space dimension $n=3072$, and the network's fully-connected architectures mentioned in Appendix \ref{appendix:architectures}, the results appear in Table \ref{table:models_weights_angle} show the average angles based on the following description:
\begin{enumerate}
\item \emph{Angles of model 1:} Average angle between $\partial \theta_1[j_1]$ and $\partial \theta_1[j_2]$ ($0\leq j_1\neq j_2<k$).
\item \emph{Angles of model 2:} Average angle between $\partial \theta'_1[j_1]$ and $\partial \theta'_1[j_2]$ ($0\leq j_1\neq j_2<k'$).
\item\emph{Angles between models:} Average angle between $\partial \theta_1[j_1]$ and $\partial \theta'_1[j_2]$ for any valid combination of indices.
\end{enumerate}

\noindent

\begin{table}[htp!]
  \centering
  \caption{The average angle between the ``vectors of updates'' ($\boldsymbol{n=3072}$).}\label{table:models_weights_angle}
  \begin{tabular}{ c c c c}
    \toprule
    Architecture & Angles of model 1  & Angles of model 2 &  Angles between models \\
    \midrule
    Shallow networks &$72.83^{\circ}$ & $74.22^{\circ}$& $73.74^{\circ}$ \\ 
    Deep networks & $75.40^{\circ}$ & $75.29^{\circ}$& $75.92^{\circ}$ \\ 
    Deep Vs Shallow & $73.45^{\circ}$ & $74.37^{\circ}$& $74.02^{\circ}$ \\ 
    \bottomrule
  \end{tabular}
\end{table}

\noindent
For the same settings, where the generated ``vectors of updates'' are correlated, training the network with a \emph{significant} learning rate (such that the initial weights are negligible compared to the derivatives) for a single step will generate new weights $\overline{\theta}_1,\overline{\theta}'_1$ that will be significantly biased toward $\partial \theta_1, \partial \theta'_1$, respectively. Table \ref{table:weights_adver_angle} shows the following:

\begin{enumerate}
\item \emph{Adversarial angles of model 1:} Average angle between $\overline{\theta}_1[j]$ ($0\leq j<k$) and the adversarial vector w.r.t the first model (average over $100$ adversarial vectors).

\item \emph{Adversarial angles of model 2:} Average angle between $\overline{\theta}'_1[j]$ (for $0\leq j <k'$) and the adversarial vector w.r.t the second model (average over $100$ adversarial vectors).
\end{enumerate}

\begin{table}[htp!]
  \centering
  \caption{The angle between the ``vectors of weights'' of each model and the adversarial vectors.}\label{table:weights_adver_angle}
  \begin{tabular}{ c c c}
    \toprule
    Architecture & Adversarial angles of model 1  &  Adversarial angles of model 2\\
    \midrule
    Shallow networks & $68.31^{\circ}$& $67.51^{\circ}$ \\ 
    Deep networks &$72.78^{\circ}$ & $72.61^{\circ}$ \\ 
    Deep Vs Shallow &$66.01^{\circ}$ & $66.9^{\circ}$ \\ 
    \bottomrule
  \end{tabular}
\end{table}

Based on Appendix \ref{appendix:angles_random_vectors}, for the input dimension of $3072$, the angles we expect from each cell in Tables \ref{table:models_weights_angle} and \ref{table:weights_adver_angle} should be $\sim 88.96^{\circ}$, and the probability of getting an angle of $\sim 70^{\circ}$ is extremely low.

\subsection{Optimizers}
\label{optimizers}
The above explanation (in this section) was demonstrated using stochastic gradient descent steps, known as ``SGD'' \cite{spall2005introduction}, where we update the weights in iteration $t$ as follows:
$$\theta^{t+1}=\theta^{t}-lr\cdot\partial \theta^{t}$$

Another commonly used optimizer is ``Adam'' \cite{kingma2017adam}, with the following iterative optimization:
$$\mathcal{V}^{t+1}=\beta_1\cdot \mathcal{V}^{t}+(1-\beta_1)\cdot\partial \theta^{t}$$
$$\mathcal{S}^{t+1}=\beta_2\cdot \mathcal{S}^{t}+(1-\beta_2)\cdot(\partial \theta^{t})^2$$
$$\mathcal{S_C}^{t+1}=\mathcal{S}^{t+1}/(1-\beta_2^{t+1}), \mathcal{V_C}^{t+1}=\mathcal{V}^{t+1}/(1-\beta_1^{t+1})$$
$$\theta^{t+1}=\theta^{t}-lr\cdot\frac{\mathcal{V_C}^{t+1}}{\sqrt{\mathcal{S_C}^{t+1}+\epsilon}}$$

For the usually predetermined constants $\beta_1=0.9,\beta_2=0.999,\epsilon=1e-8$, where the standard initial conditions are $\mathcal{V}^0=0,\mathcal{S}^0=0,t=0$, we get that the first step results in the following expression:

$$\mathcal{V}^1=0.1\partial \theta^0, \mathcal{S}^1=0.001(\partial \theta^0)^2$$
$$\mathcal{V_C}^1=\partial \theta^0, \mathcal{S_C}^1=(\partial \theta^0)^2$$
$$\theta^1=\theta^0-lr\frac{\partial \theta^0}{\sqrt{(\partial \theta^0)^2+\epsilon}}\sim \theta^0-lr\frac{\partial \theta^0}{|\partial \theta^0|}$$

We get that the expression $\frac{\partial \theta^0}{|\partial \theta^0|}$ (up to the small error generated by $\epsilon$) is a vector with the elements ${\pm 1,0}$ (independently of the metric we are using - $\ell_0,\ell_1,\ell_2,\ell_{\infty}$). The earlier the step is, the closer expression to $\frac{\partial \theta^0}{|\partial \theta^0|}$ we'll get.

This occurrence leads to a reasonable explanation for the emphasized effect of early transferability using Adam optimizer. The weights' update for Adam is $lr\frac{\partial \theta^0}{|\partial \theta^0|}$, where the stochastic gradient step is $lr \cdot \partial \theta^0$. Typically, the values of $|\partial \theta^0|$ are much smaller than $1$ in the traditional SGD. Therefore we get (excluding zero elements):
$$\bigg|lr\frac{\partial \theta^0}{|\partial \theta^0|}\bigg|>\bigg|lr\partial \theta^0\bigg|$$
Overall the update using Adam optimizer is much more ``aggressive''  than the update of SGD. Since early transferability depends on the values of the derivatives of the weights, a larger step will emphasize the effect. Therefore Adam optimizer could use a lower learning rate in order to create the same effect compared to SGD.

\subsection{The influence on the general training process}
\label{long_term}
According to the above analysis (section \ref{Occurrence explanation}), a massive influence of the weights' update might create a noticeable occurrence of early transferability. The general idea of early transferability concerns the weights' update of the two given networks. We experimentally found that the vectors of updates are unexpectedly correlated. These correlated vectors of updates might lead the state of the two networks to be very close to the same local minima after very few steps of the training process. It might increase the probability of the networks converging to a similar solution in the long-term perspective of the training. Therefore, we might suspect that early transferability might influence long-term training. In particular, a prominent occurrence of early transferability in the initial steps might increase the long-term transferability.

We present a similar analysis to the one described in section \ref{experimental_results}, but in the current analysis, we desire to capture the long-term properties of the training process. Therefore, we measure the horizontal timeline of the training with epochs rather than with steps. Moreover, we want to present the long-term evolution using two settings. One of the settings has a significant occurrence of the early transferability effect (observed in the early stages). The other has a low occurrence of the effect. To avoid any other factor that might influence the experiment, we use similar parameters for both cases, except for one parameter, the learning rate.

\begin{figure}[htp!]
     \includegraphics[width=1\textwidth]{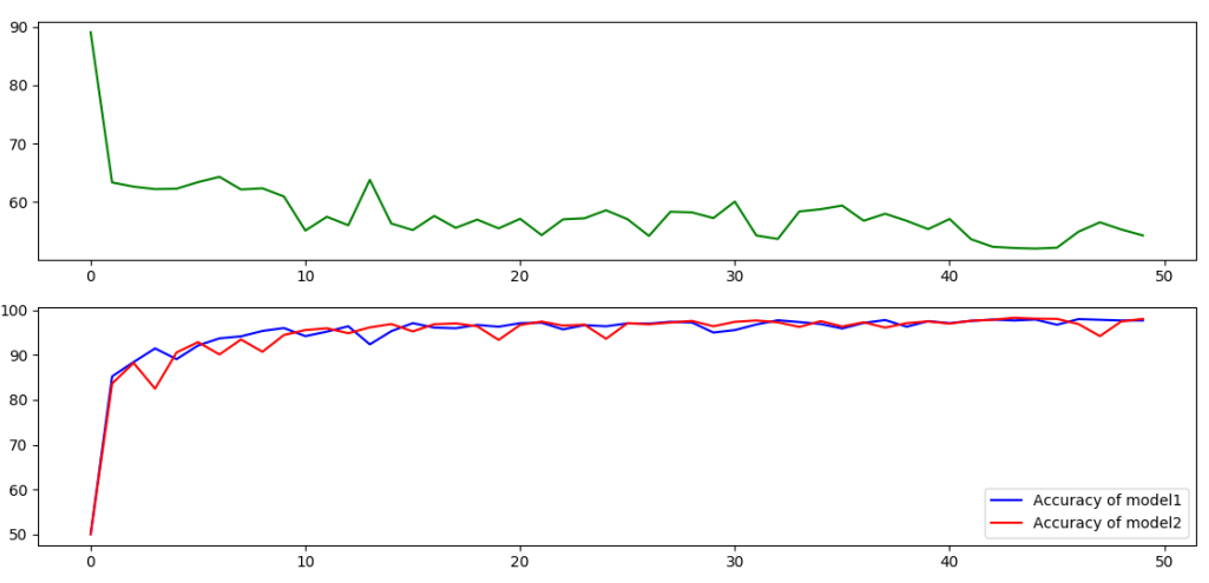}
     \caption{Long-term influence using epochs timeline. Learning rate of $1e-2$}
     \label{fig:long_term_1e2}
\end{figure}

\begin{figure}[htp!]
     \includegraphics[width=1\textwidth]{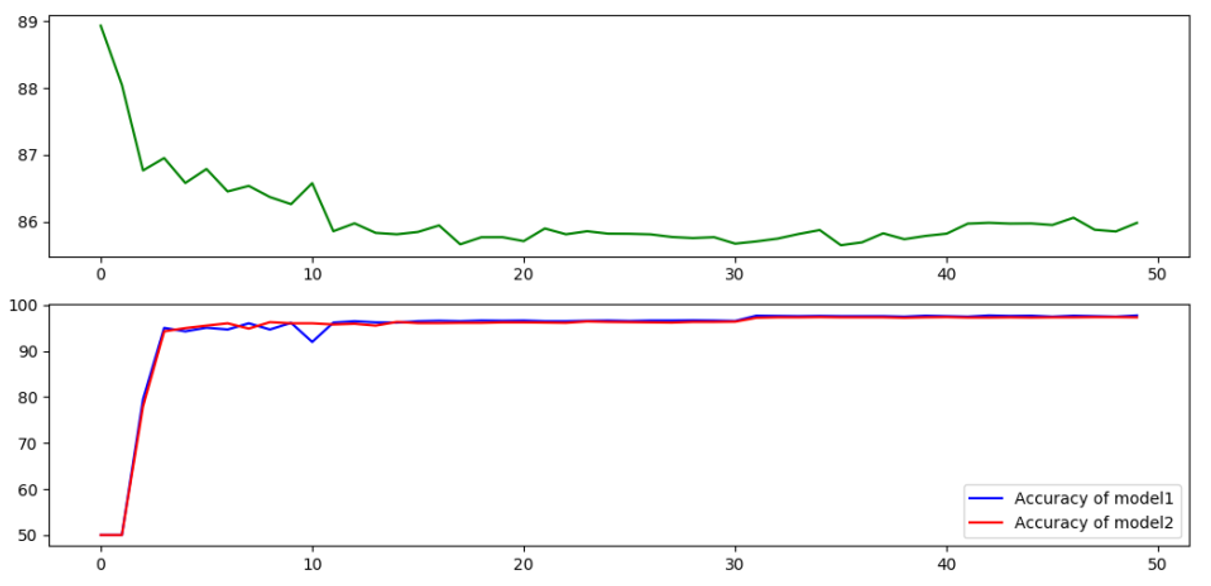}
     \caption{Long-term influence using epochs timeline. Learning rate of $1e-4$}
     \label{fig:long_term_1e4}
\end{figure}

In Figures \ref{fig:long_term_1e2} and \ref{fig:long_term_1e4}, we use a batch size of $128$, Adam optimizer, and CIFAR10 dataset \cite{CIFAR10} with the categories \emph{Airplane} and \emph{Automobile}. We use Architecture B, which appears in Appendix \ref{appendix:architectures}. The learning rates used in Figures \ref{fig:long_term_1e2} and \ref{fig:long_term_1e4} are $1e-2,1e-4$, respectively. We can clearly see that both models converged and are in a steady-state. However, the top graph presented in both figures points to a massive difference in the angle between the perturbations, while the bottom graphs in both figures show approximately the same results. The difference between the long-term transferability occurrences might be explained by the different impacts of the early transferability effects during the initial stages of the training.

Additional experiments appear in Appendix \ref{appendix:additional_results}.

\section{Limitations}
Early transferability might occur for models trained with reasonable hyperparameters, as demonstrated in this paper. Yet, the consequences might change for different tuning. In particular, for optimizers such as SGD, the effect might occur only for a significant  learning rate as opposed to Adam optimizer.
Additionally, the effect occurs in the best manner when the hyperparameters of the reference and target model are similar (even though identical parameters are not required).

\section{Conclusion}
Our findings provide evidence for an unknown and counter-intuitive phenomenon. One of the key insights behind this is the unexpectedly early stages, in which the adversarial directions start to align with each other. We have conducted a supportive experimental analysis for several architectures (fully-connected, convolutional, residual), datasets (CIFAR10, ImageNet), and many hyperparameters. Our findings prompt us to view transferability as a combined occurrence of ``early transferability'' and ``traditional transferability'', where the first one could be reduced or emphasized by different (but reasonable) hyperparameters. In addition, we explain why some optimizers might generate a greater impact than others. Our experimental results and proposed explanations shed light on various factors such as the possible long-term influence of ``early transferability'' during the training process and the unrelated impact of this effect w.r.t. the data complexity, or the ability of the network to learn.
In section \ref{goal}, we present the essence of this effect and hope that the discovered occurrence will reduce the mysteries surrounding ``Transferability''.

\bibliography{main}

\appendix

\section{Angles between random vectors}
\label{appendix:angles_random_vectors}
Consider a vector space $V=\mathbb{R}^n$ and two random vectors $a=(a_1,...,a_n)\in\mathbb{R}^n$ and $b=(b_1,...,b_n)\in\mathbb{R}^n$ sampled independently from the unit sphere's surface. The process of choosing the vectors is invariant to rotations of the axes system. Therefore it's equivalent to representing $a$ and $b$ as $a=(1,0,...,0)$ and $b=(b_1',...,b_n')$.

\noindent
The projection of $b$ onto $a$ is:
$$proj_ab=\frac{a\cdot b}{\|a\|}=\frac{a\cdot b}{1}=a\cdot b=1\cdot b_1' + \sum_{i=2}^n{0\cdot b_i'}=b_1'$$

\begin{mdframed}[linewidth=2pt,backgroundcolor=yellow!20,userdefinedwidth=120pt,align=center]
$\boldsymbol{proj_ab=\frac{a\cdot b}{\|a\|}=b_1'}$
\end{mdframed}

\noindent
Due to symmetry, we get that $\forall_{1\leq i,j\leq n}:\mathbb{E}(|b_i'|^p)=\mathbb{E}(|b_j'|^p)$. Additionally, for $\ell_p$ norm, we know that $\|b\|=\big({\sum_{i=1}^n{|b_i'|^p}\big)^{1/p}}=1$, therefore:
$$1=
\mathbb{E}(\sum_{i=1}^n{|b_i'|^p})=
\sum_{i=1}^n{\mathbb{E}(|b_i'|^p)}=
\sum_{i=1}^n{\mathbb{E}(|b_i'|^p)}=
n\cdot\mathbb{E}(|b_1'|^p)$$

\begin{mdframed}[linewidth=2pt,backgroundcolor=yellow!20,userdefinedwidth=90pt,align=center]
$\boldsymbol{\mathbb{E}(|b_1'|^p)=\frac{1}{n}}$
\end{mdframed}

\noindent
The angle $\theta$ between $a$ and $b$ satisfies the following:
$$\frac{a\cdot b}{\|a\|\|b\|}=\cos{\theta}$$
Since we already calculated $\frac{a\cdot b}{\|a\|}$ and we know that $\|b\|=1$, we get:
$$\cos{\theta}=\frac{a\cdot b}{\|a\|\|b\|}=\frac{a\cdot b}{\|a\|}\cdot\frac{1}{\|b\|}=b_1'\cdot\frac{1}{1}=b_1'$$

\noindent
The expected value of $\cos^p{\theta}$ is $\mathbb{E}(|b_1'|^p)=\frac{1}{n}$. Simplifying this expression will produce an average angle of $\theta=\arccos{\big((\frac{1}{n})^{1/p}\big)}$ for positive values of $b_1'$ and $\theta=\arccos{\big(-(\frac{1}{n})^{1/p}\big)}$ for negative values of $b_1'$.

\begin{mdframed}[linewidth=2pt,backgroundcolor=yellow!20,userdefinedwidth=130pt,align=center]
$\boldsymbol{\theta=\arccos{\big((\frac{1}{n})^{1/p}\big)}}$\\
$\boldsymbol{\theta=\arccos{\big(-(\frac{1}{n})^{1/p}\big)}}$
\end{mdframed}

\noindent
For positive correlation between $a$ and $b$ and $\ell_2$ norm (which produces an average angle of $\theta=\arccos{\frac{1}{\sqrt{n}}}$), we will calculate some examples for different space dimensions and their expected angle in Table \ref{table:expected_angles_high_dim}.

\begin{table}[htp!]
  \centering
  \caption{Examples for different space dimensions and the expected angle between random vectors in that dimension (for positive correlation and $\ell_2$ norm).}\label{table:expected_angles_high_dim}
  \begin{tabular}{ c c}
    \toprule
    Dimension($n$) & Expected angle\\
    \midrule
    $2$ &  $45.0^{\circ}$ \\
    $16$ &  $75.52^{\circ}$ \\
    $128$ &  $84.92^{\circ}$ \\
    $784$ &  $87.92^{\circ}$ \\
    $3072$ &  $88.96^{\circ}$ \\
    $196608$ &  $89.87^{\circ}$ \\
    \bottomrule
  \end{tabular}
\end{table}

\noindent
Additionally, the variance is:
$$\mathbb{V}(b_1')=\mathbb{E}(b_1'^2)-\mathbb{E}(b_1')^2=
\mathbb{E}(b_1'^2)-0=\mathbb{E}(|b_1'|^2)=\frac{1}{n}$$
therefore, for higher dimensions we get smaller variance. Based on that, the probability of getting small angles between random vectors in high dimensions is much lower.

\noindent
Markov's inequality states the following:
$$P(X\geq a)\leq\frac{\mathbb{E}(X)}{a}$$
for non-negative random variable $X$ and $a>0$.

\noindent
We can use Markov's inequality to upper bound the probability of getting some angle $\theta$ between random vectors. We will use the following input $X=|b_1'|^2, a=\frac{t}{n}$ for integer $t>0$ and $\mathbb{E}(|b_1'|^2)=\frac{1}{n}$:
$$P(|b_1'|^2\geq\frac{t}{n})\leq
\frac{\mathbb{E}(|b_1'|^2)}{a}=
\frac{\frac{1}{n}}{\frac{t}{n}}=\frac{1}{t}
$$

\noindent
It means that the probability of getting $\theta\leq\arccos{\big((\frac{t}{n})^{1/2}\big)}$ is lower than $\frac{1}{t}$. In Table \ref{table:angles_probabilities} we can see the probability of getting different angles for dimension size of $3072$.

\begin{table}[htp!]
  \centering
  \caption{An upper bound for the probability of getting angles between random vectors for $n=3072$, based on different $t$ values.}\label{table:angles_probabilities}
  \begin{tabular}{ c c c}
    \toprule
    Assigned $\boldsymbol{t}$& Angle & Probability's upper bound ($\boldsymbol{1/t}$)\\
    \midrule
    $1$ &  $89.96^{\circ}$ & $1.0$ \\
    $2$ &  $88.53^{\circ}$ & $0.5$\\
    $10$ &  $86.72^{\circ}$ & $0.1$\\
    $100$ &  $79.60^{\circ}$ & $0.01$ \\
    $350$ &  $70.27^{\circ}$ & $0.0028$ \\
    $1000$ &  $55.21^{\circ}$ & $0.001$ \\
    \bottomrule
  \end{tabular}
\end{table}

In Table \ref{table:angles_probabilities}, we concluded only an upper bound for the discussed probabilities, in reality the probabilities are even lower. For example, we can experimentally calculate the angles for a large number of $100,000$ pairs of random vectors of size $3072$. The \emph{minimum} angle we are getting out of the $100,000$ angles is $85.29^{\circ}$.

\section{Architectures}
\label{appendix:architectures}
\subsection{Fully-connected architectures}
\begin{figure}[htp!]
\centering
\includegraphics[width=12cm]{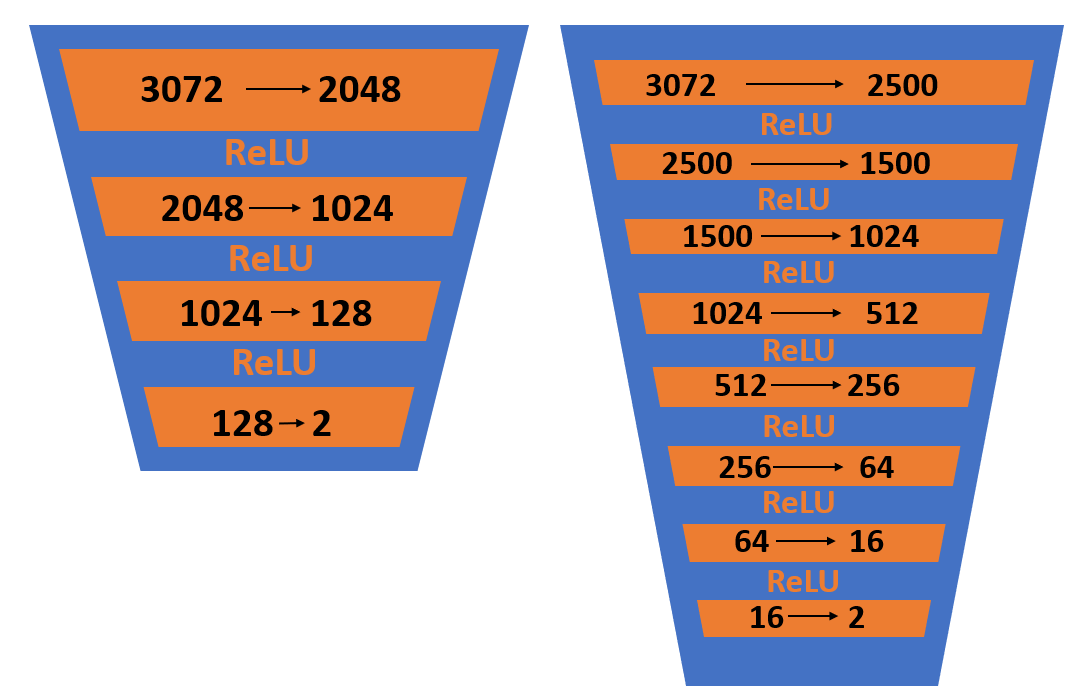}
\caption{The architectures of the ``Shallow network'' (on the left) and the ``Deep network'' (on the right).}
\label{fig:networks}
\end{figure}

In Figure \ref{fig:networks}, we can see the two fully-connected architectures used in this work. We address them as the ``Shallow network'' and the ``Deep network''.

\subsection{Convolutional architectures}
\begin{description}
\item[Architecture A:] A shallow architecture of a convolutional network:
\begin{itemize}
    \item \textbf{Convolutional layer} [input filters: 3, output filters: 128]
        \subitem kernel size: 5, stride: 2, padding: 1
    \item ReLU layer
    \item\textbf{ Convolutional layer} [input filters: 128, output filters: 128]
        \subitem kernel size: 5, stride: 2, padding: 1
    \item ReLU layer
    \item \textbf{Convolutional layer} [input filters: 128, output filters: 256]
        \subitem kernel size: 5, stride: 2, padding: 1
    \item ReLU layer
    \item \textbf{Convolutional layer} [input filters: 256, output filters: 256]
        \subitem kernel size: 5, stride: 2, padding: 1
    \item \textbf{ReLU layer}
    \item \textbf{Flatten}
    \item \textbf{Fully-connected layer} [input: 256, output: 2]
\end{itemize}

\item[Architecture B:] A deep architecture of a convolutional network:
\begin{itemize}
    \item \textbf{Convolutional layer} [input filters: 3, output filters: 128]
        \subitem kernel size: 3, stride: 1, padding: 1
    \item \textbf{BatchNorm layer} [filters: 128]
    \item \textbf{ReLU layer}
    \item\textbf{ Convolutional layer} [input filters: 128, output filters: 128]
        \subitem kernel size: 3, stride: 1, padding: 1
    \item \textbf{BatchNorm layer} [filters: 128]
    \item \textbf{ReLU layer}
    \item \textbf{MaxPool layer}
        \subitem kernel size: 2, stride: 2, padding: 0, dilation: 0
    
    \item \textbf{Convolutional layer} [input filters: 128, output filters: 256]
        \subitem kernel size: 3, stride: 1, padding: 1
    \item \textbf{BatchNorm layer} [filters: 256]
    \item \textbf{ReLU layer}
    \item\textbf{ Convolutional layer} [input filters: 256, output filters: 256]
        \subitem kernel size: 3, stride: 1, padding: 1
    \item \textbf{BatchNorm layer} [filters: 256]
    \item \textbf{ReLU layer}
    \item \textbf{MaxPool layer}
        \subitem kernel size: 2, stride: 2, padding: 0, dilation: 0
    
    \item \textbf{Convolutional layer} [input filters: 256, output filters: 512]
        \subitem kernel size: 3, stride: 1, padding: 1
    \item \textbf{BatchNorm layer} [filters: 512]
    \item \textbf{ReLU layer}
    \item\textbf{ Convolutional layer} [input filters: 512, output filters: 512]
        \subitem kernel size: 3, stride: 1, padding: 1
    \item \textbf{BatchNorm layer} [filters: 512]
    \item \textbf{ReLU layer}
    \item \textbf{MaxPool layer}
        \subitem kernel size: 2, stride: 2, padding: 0, dilation: 0
        
    \item\textbf{ Convolutional layer} [input filters: 512, output filters: 1024]
        \subitem kernel size: 3, stride: 1, padding: 0
    \item \textbf{BatchNorm layer} [filters: 1024]
    \item \textbf{ReLU layer}
    \item \textbf{MaxPool layer}
        \subitem kernel size: 2, stride: 2, padding: 0, dilation: 0
        
    \item \textbf{Flatten}
    \item \textbf{Fully-connected layer} [input: 1024, output: 2]
\end{itemize}

\end{description}

\section{Additional experiments}
\label{appendix:additional_results}
\subsection{Details}
The following experiments demonstrate many cases that clearly show the early transferability effect. The settings appear in Table \ref{table:parmeters_combinations}. The architectures are based on Appendix \ref{appendix:architectures} and Resnet18 \cite{he2016deep}. Cases 1-8 consider the categories \emph{Cat} and \emph{Dog} of the dataset CIFAR10. Case 9 address the categories \emph{Goldfish} and \emph{White shark} of the dataset ImageNet.

\begin{table}[htp!]
  \centering
  \caption{Possible parameters we will use in the experiments.}\label{table:parmeters_combinations}
  \begin{tabular}{ c c c c c}
    \toprule
    & Model 1 & Model 2 & Learning rate & Batch size\\
    \midrule
    Case 1 & FC Deep network & FC Deep network &  1e-3 &  128 \\ 
    Case 2 & FC Deep network & FC Deep network &  1e-4 &  1024 \\ 
    Case 3 & FC Shallow network & FC Shallow network &  1e-2 &  128 \\ 
    Case 4 & FC Shallow network & FC Deep network &  1e-2 &  128 \\ 
    Case 5 & FC Shallow network & FC Deep network &  1e-3 &  1024 \\ 
    Case 6 & Convolutional Architecture A & Convolutional Architecture A &  1e-3 &  1024 \\ 
    Case 7 & Convolutional Architecture B & Convolutional Architecture A &  1e-3 &  128 \\ 
    Case 8 & Resnet18 & Resnet18 &  1e-3 &  16 \\ 
    \bottomrule
  \end{tabular}
\end{table}

\subsection{results}
\begin{figure}[htp!]
     \includegraphics[width=1\textwidth]{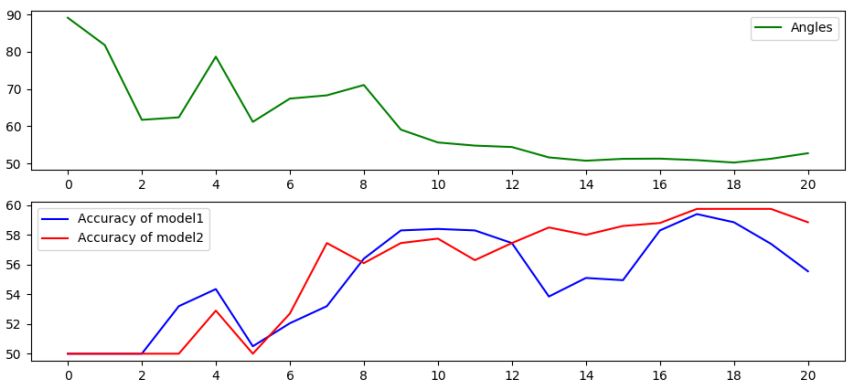}
     \caption{Early transferability - Case 1.}
     \label{fig:case1}
\end{figure}

\begin{figure}[htp!]
     \includegraphics[width=1\textwidth]{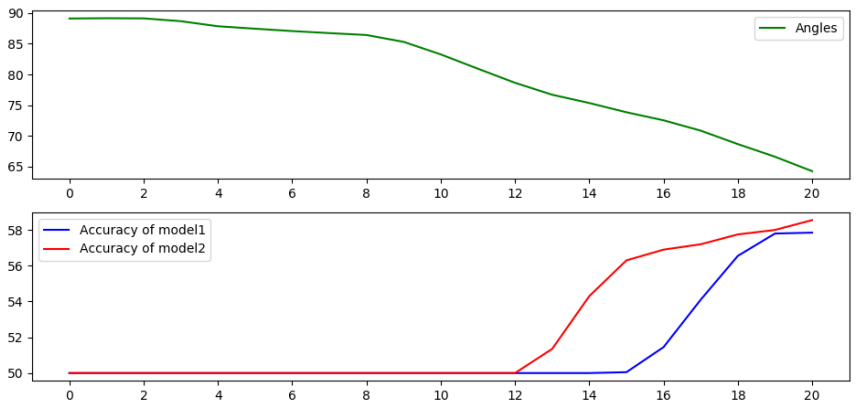}
     \caption{Early transferability - Case 2.}
     \label{fig:case2}
\end{figure}

\begin{figure}[htp!]
     \includegraphics[width=1\textwidth]{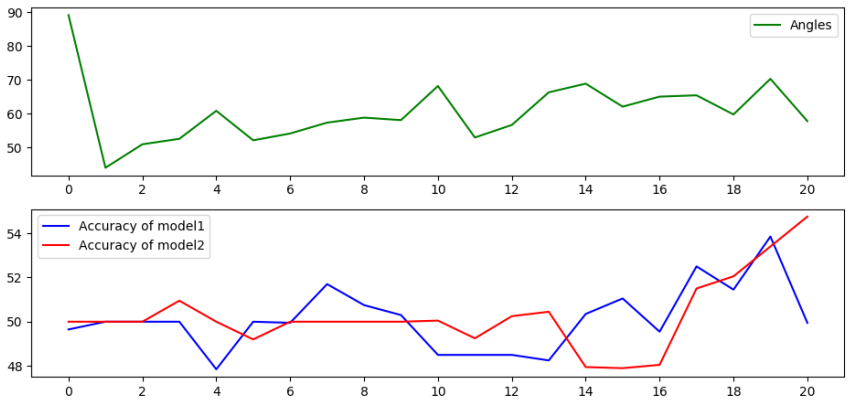}
     \caption{Early transferability - Case 3.}
     \label{fig:case3}
\end{figure}

\begin{figure}[htp!]
     \includegraphics[width=1\textwidth]{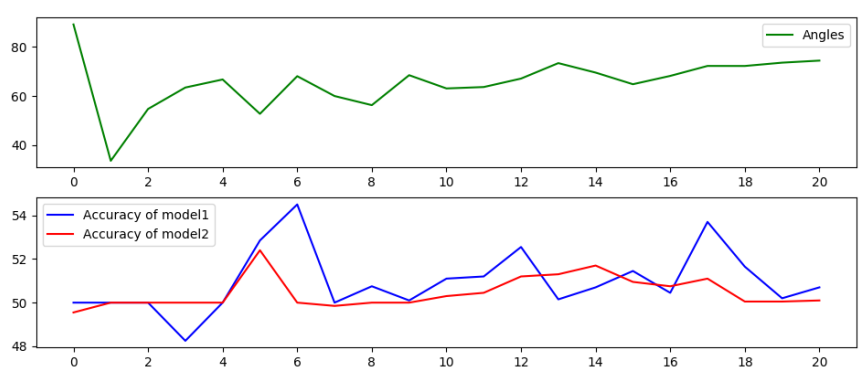}
     \caption{Early transferability - Case 4.}
     \label{fig:case4}
\end{figure}

\begin{figure}[htp!]
     \includegraphics[width=1\textwidth]{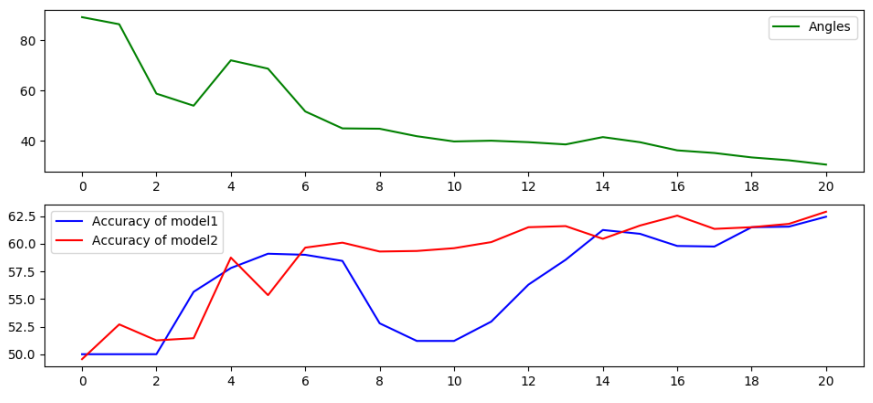}
     \caption{Early transferability - Case 5.}
     \label{fig:case5}
\end{figure}

\begin{figure}[htp!]
     \includegraphics[width=1\textwidth]{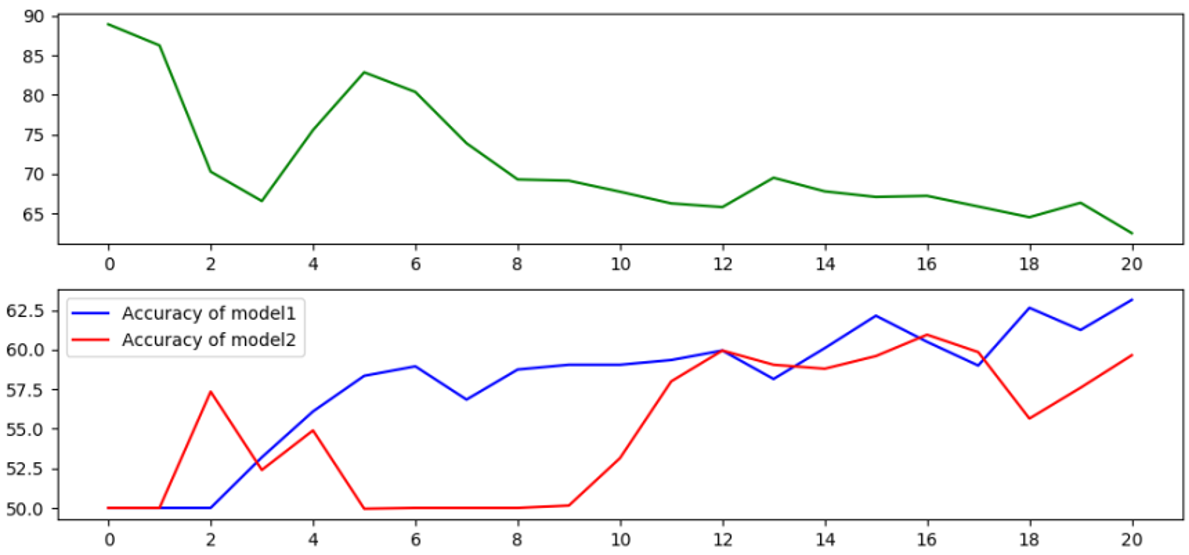}
     \caption{Early transferability - Case 6.}
     \label{fig:case6}
\end{figure}

\begin{figure}[htp!]
     \includegraphics[width=1\textwidth]{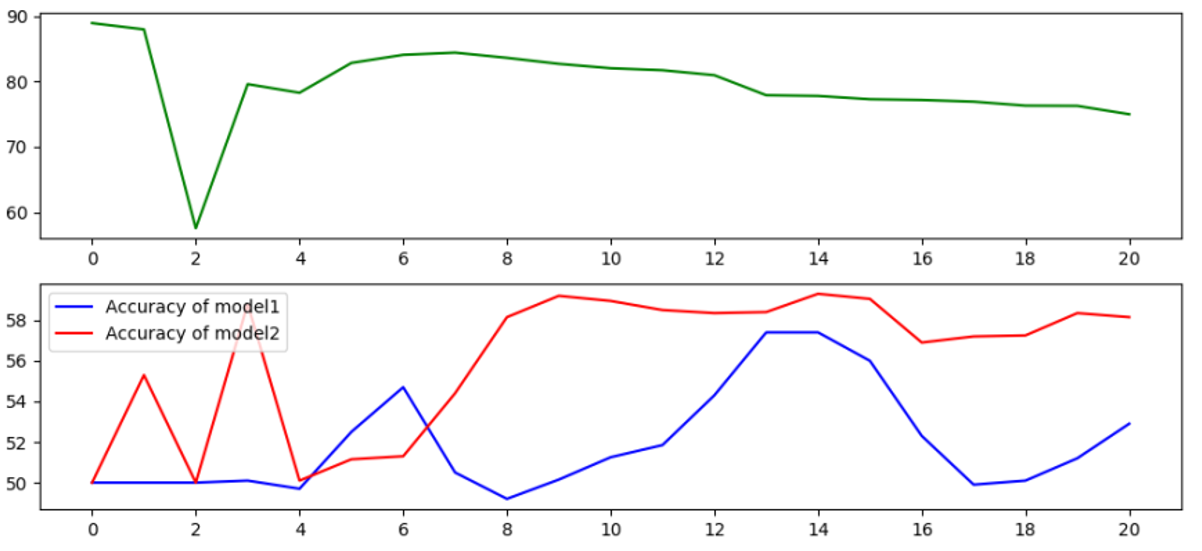}
     \caption{Early transferability - Case 7.}
     \label{fig:case7}
\end{figure}

\begin{figure}[htp!]
     \includegraphics[width=1\textwidth]{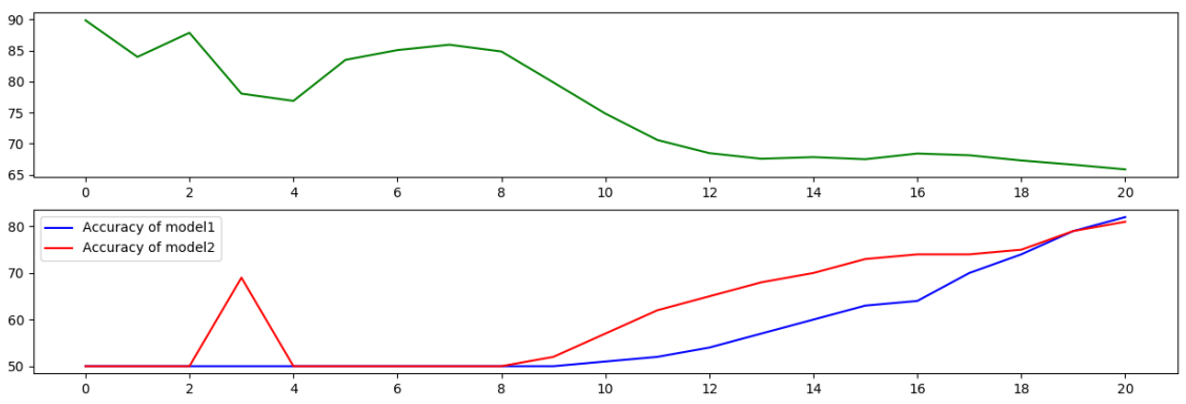}
     \caption{Early transferability - Case 8.}
     \label{fig:case8}
\end{figure}

In Figures \ref{fig:case1},\ref{fig:case2},\ref{fig:case3},\ref{fig:case4},\ref{fig:case5},\ref{fig:case6},\ref{fig:case7},\ref{fig:case8}, we can see the graph analysis that was mentioned in the section \ref{experimental_results} of all the cases in Table \ref{table:parmeters_combinations}.

\subsection{long-term transferability}
\begin{figure}[htp!]
     \includegraphics[width=1\textwidth]{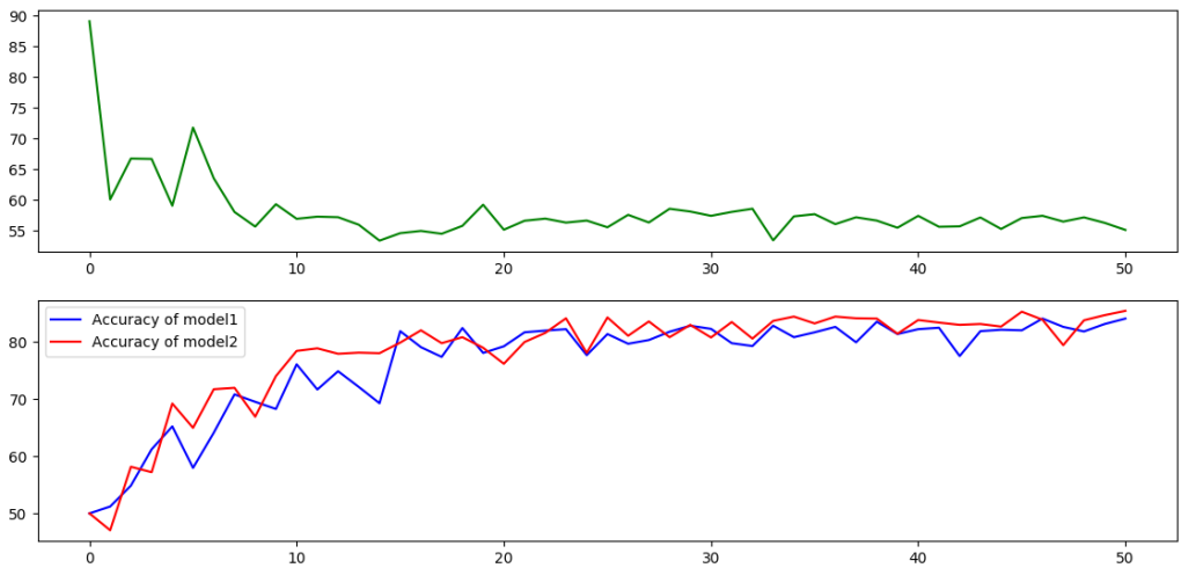}
     \caption{Long-term influence using epochs timeline. Learning rate of $1e-2$}
     \label{fig:additional_long_term_1e2}
\end{figure}

\begin{figure}[htp!]
     \includegraphics[width=1\textwidth]{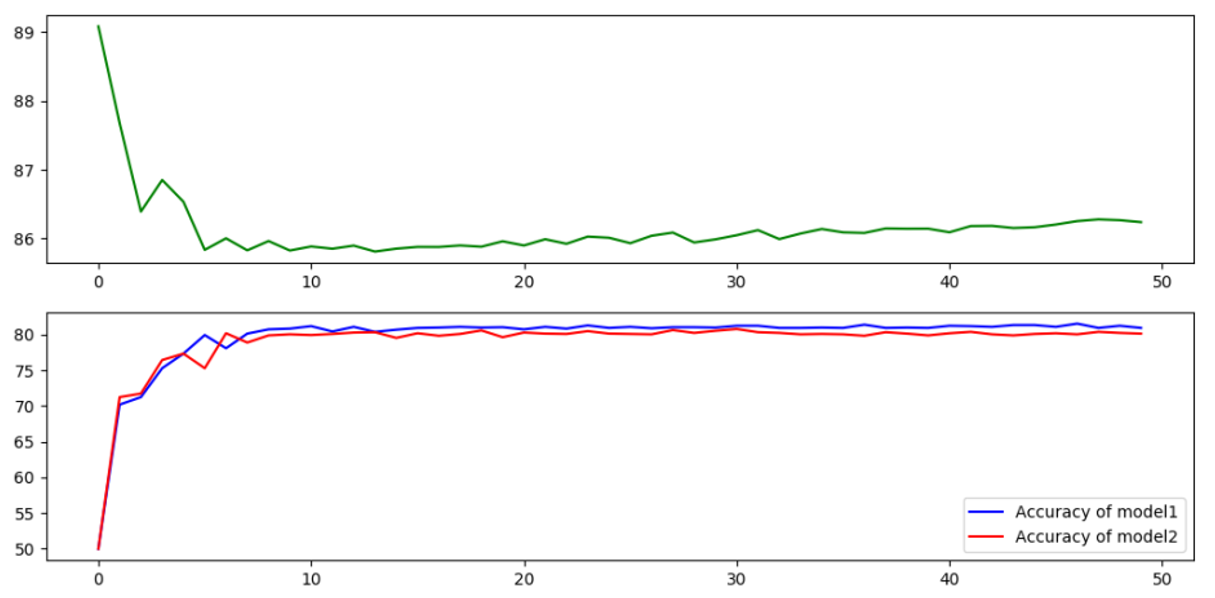}
     \caption{Long-term influence using epochs timeline. Learning rate of $1e-4$}
     \label{fig:additional_long_term_1e4}
\end{figure}

Following section \ref{long_term}, we conduct an additional experiment using similar settings but with a different pair of categories of the dataset CIFAR10, \emph{Cat} and \emph{Dog}. In Figures \ref{fig:additional_long_term_1e2} and \ref{fig:additional_long_term_1e4}, we can see a similar occurrence to the one we witnessed in section \ref{long_term}.

\end{document}